\documentclass[11pt]{article}

\usepackage[final]{acl}

\usepackage{times}
\usepackage{latexsym}
\usepackage[T1]{fontenc}
\usepackage[utf8]{inputenc}
\usepackage{microtype}
\usepackage{inconsolata}
\usepackage{graphicx}
\usepackage{booktabs}
\usepackage{amsmath}
\usepackage{multirow}
\usepackage{makecell}
\usepackage{array}
\usepackage{xcolor}
\usepackage{tikz}
\usetikzlibrary{shapes,arrows.meta,positioning,fit,backgrounds,decorations.pathreplacing}
\usepackage[skins,breakable]{tcolorbox}
\usepackage{balance}  

\definecolor{clrSFT}{HTML}{E8A087}    
\definecolor{clrCls}{HTML}{88C4C8}    
\definecolor{clrThird}{HTML}{B0CA97}  
\definecolor{clrEns}{HTML}{D1BC8A}    

\newcommand{\fone}{\mathit{F1}}
\newcommand{\fcv}{\mathit{F1}_{\mathrm{cv}}}
\newcommand{\fcvtop}{\mathit{F1}_{\mathrm{cv}}^{\mathrm{top3}}}
\newcommand{\ftest}{\mathit{F1}_{\mathrm{test}}}

\tcbset{promptbox/.style={
  colback=clrThird!8!white, colframe=clrThird!70!black,
  fontupper=\fontsize{7.5}{9}\selectfont, boxrule=0.5pt,
  left=4pt, right=4pt, top=3pt, bottom=3pt,
  toptitle=1.5pt, bottomtitle=1.5pt,
}}

\title{N\"urnberg NLP at ChildSafeAds 2026:\\Structurally Dissimilar Voter Ensembles under Four Levels of Data Access}

\author{Philipp Steigerwald \quad Eric Rudolph \quad Jens Albrecht \\
  Technische Hochschule N\"urnberg Georg Simon Ohm \\
  \texttt{\{philipp.steigerwald,\,eric.rudolph,\,jens.albrecht\}@th-nuernberg.de}}

\begin{document}
\maketitle

\begin{abstract}
We describe the N\"urnberg NLP system for ChildSafeAds 2026.
The shared task asks what a monitoring system for commercial content in child-facing YouTube videos can achieve at a given level of data access.
We answer with per-subtask ensembles of nine voters, organised into three branches that differ in backbone, adaptation method and class scope.
Selection rests on channel-disjoint cross-validation, with the development set as a transfer check.
\textbf{The system wins two of the three subtasks}.
Its product-category score (\textbf{ST2, $\boldsymbol{0.8243}$}) and its compliance-flag score (\textbf{ST3, $\boldsymbol{0.6530}$}) are the best of the 22 final entries, and it places third on the task mean ($0.7079$).
We further compare four access levels and report the cost at test-set scale.
\end{abstract}

\section{Introduction}
\label{sec:intro}
Advertising directed at children is regulated more strictly than advertising in general, yet sponsored segments in child-facing YouTube videos are barely monitored.
The ChildSafeAds shared task \citep{bertaglia2026childsafeads} frames this as a measurement problem.
A system receives a sponsored segment together with a chosen level of data access.
It labels the commercial type of the segment (ST1), the product category (ST2) and EU consumer-law compliance risk flags (ST3).
The inputs range from the video transcript alone (L1) through video context (L2) and the channel name (L3) to the promoted product's web page (L4).
Each subtask is scored by macro-$\fone$ over the classes occurring in the reference labels.
The official ranking uses the mean of the three subtask scores.
The hidden test channels are disjoint from the training channels.

\begin{figure}[t!]
\centering
\resizebox{\columnwidth}{!}{%
\begin{tikzpicture}[
    sysbox/.style={rectangle, rounded corners=3pt, dashed, line width=0.9pt,
        minimum width=2.46cm, minimum height=1.58cm, align=center,
        inner ysep=2pt, inner xsep=3pt},
    voterchip/.style={rectangle, rounded corners=1.5pt,
        dash pattern=on 1.2pt off 1.2pt, line width=0.55pt,
        minimum width=0.70cm, minimum height=0.40cm, align=center,
        inner sep=1pt, font=\fontsize{6.3}{7.6}\selectfont},
    major/.style={rectangle, rounded corners=3pt, line width=0.9pt,
        align=center, inner ysep=4pt, inner xsep=4pt},
    edge/.style={-{Stealth[length=4pt]}, line width=0.6pt, black!50},
    ttl/.style={font=\fontsize{7.5}{9}\selectfont\bfseries},
    sub/.style={font=\fontsize{6}{7.2}\selectfont, text=black!75},
    leg/.style={anchor=west, font=\fontsize{6.3}{7.6}\selectfont, text=black!60},
]
\node[sysbox, draw=clrCls!70!black, fill=clrCls!22] at (-2.6, 0.79) {};
\node[sysbox, draw=clrThird!72!black, fill=clrThird!30] at (0, 0.79) {};
\node[sysbox, draw=clrSFT!70!black, fill=clrSFT!22] at (2.6, 0.79) {};
\node[ttl] at (-2.6, 1.30) {Generalist (G)};
\node[ttl] at (0, 1.30) {Specialist (S)};
\node[ttl] at (2.6, 1.30) {G / S / MCS};
\node[sub] at (-2.6, 1.02) {trained on all subtasks};
\node[sub] at (0, 1.02) {trained on one subtask};
\node[sub] at (2.6, 1.02) {submission-dependent};
\foreach \xc/\clr in {-2.6/clrCls, 0/clrThird, 2.6/clrSFT}{
  \foreach \dx in {-0.79, 0, 0.79}{
    \node[voterchip, draw=\clr!60!black, fill=white!45!\clr!14]
      at ({\xc+\dx}, 0.44) {voter};
    \draw[edge] ({\xc+\dx}, 0.23) -- ({\xc+\dx}, -0.30);
  }
}
\node[major, draw=clrEns!80!black, fill=clrEns!28,
      minimum width=7.66cm, minimum height=0.74cm,
      font=\fontsize{6.8}{8.2}\selectfont] (ens) at (0, -0.68)
    {$\widehat{Y}_t=P_t\!\left(\left\{\ell\in\mathcal{L}_t\mid
      \textstyle\sum_{j=1}^{9}\mathbf{1}[\ell\in v_j^{(t)}]\geq5\right\}\right),
      \quad t=1,2,3$};
\node[rectangle, rounded corners=1.5pt, draw=black!55, dashed, line width=0.7pt,
      minimum width=0.55cm, minimum height=0.28cm] at (-2.05, -1.34) {};
\node[leg] at (-1.73, -1.34) {= one branch};
\node[rectangle, rounded corners=1.5pt, draw=black!55,
      dash pattern=on 1.2pt off 1.2pt, line width=0.5pt,
      minimum width=0.55cm, minimum height=0.28cm] at (0.45, -1.34) {};
\node[leg] at (0.77, -1.34) {= one trained model};
\end{tikzpicture}%
}
\caption{Three structurally dissimilar branches contribute three voters each to a per-subtask ensemble. Deployed ensembles vary (Table~\ref{tab:submissions}).}
\label{fig:arch}
\end{figure}
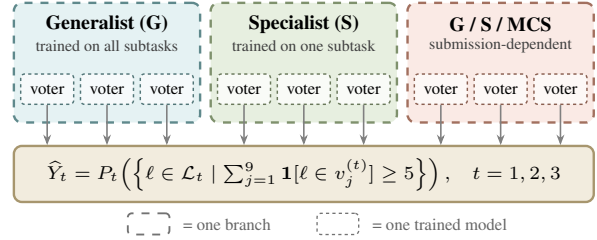

Our system is a per-subtask ensemble of nine voters (Figure~\ref{fig:arch}).
It combines structurally dissimilar voters through majority voting \citep{dietterich2000ensemble} and uses cross-validation folds as both voter pool and internal estimate \citep{krogh1995ensembles}.
A \textit{configuration}, one backbone trained by one method for one class scope at one access level, becomes a branch when deployed on its three best cross-validation folds, and each fold-model casts one vote.
The branches also differ in class scope, from a generalist (G) trained on all three subtasks through a specialist (S) for one subtask to a minority-class specialist (MCS) for the minority labels alone.
Such an ensemble approach has already proven itself in other text classification domains \citep{steigerwald2026germeval,steigerwald2026psydefdetect}.
Here it meets multi-label subtasks and a data-access dimension, and the voter pool gains an encoder branch.

On the final leaderboard the system wins two of the three subtasks.
Its ST2 ($0.8243$) and ST3 ($0.6530$) scores are the best of all 22 entries.
Our entry also leads the auxiliary ST3-family score ($0.7031$), and our best upload reached $0.7281$ on that score.
That score regroups the eight flags into four families (disclosure, content, product, housekeeping) to even out their imbalance and sits outside the ranking.
On the official mean of the three subtask scores, the system places third ($0.7079$).
The gap to the winning entry lies entirely in ST1 (\S\ref{sec:gap}).

Our contributions are (i) the leaderboard-best ST2 and ST3 systems, (ii) an aggregated voter-level comparison across four access levels, cross-checked against the deployed ensemble, and (iii) a training-and-inference cost account of the default nine-voter architecture.

\section{Related Work}
\label{sec:related}
Computational work on sponsored content has mostly studied disclosure.
About $10\%$ of affiliate content on YouTube and Pinterest carries any disclosure \citep{mathur2018endorsements}, an estimated $17.7\%$ of studied influencers' posts without a sponsorship declaration are in fact sponsored \citep{zarei2020characterising}, and undisclosed sponsorship can be ranked from text, images and influencer--brand relations \citep{kim2021discovering}.
Recent work moves from detecting sponsorship to assessing its compliance \citep{gui2024across,bertaglia2025influencer}.
Closest to this task, \citet{gui2025evaluating} prompt LLMs to decide whether a post is advertising and to justify the decision under advertising law, and report a marked drop on ambiguous posts and the most misidentified cues on hidden advertising.
ChildSafeAds instead rests on crowd-sourced segment identification, and its input is a spoken transcript with metadata rather than a caption.
Work on minors has concentrated on harmful rather than commercial content \citep{papadamou2020disturbed}, and children's exposure to advertising on video platforms is documented mainly by measurement studies \citep{khan2024analyzing,potvinkent2024junkfluenced}.
Platform self-declaration is no substitute, as channels sharing inappropriate child-directed content are less likely to set the \texttt{madeForKids} flag \citep{gkolemi2022youtubers}.
The closest methodological precedent is the CLAUDETTE line on unfair terms-of-service clauses \citep{lippi2019claudette,drawzeski2021corpus}, with comparable label sets for cookie banners \citep{vanhofslot2022automatic} and dark patterns \citep{mathur2019dark}.
The terms-of-service and cookie-banner work operationalises a legal instrument as a small expert-designed multi-label taxonomy.
All of these works share the caveat that an accurate classifier of such labels is not thereby a detector of the legal wrong \citep{soe2022automated,santosh2022deconfounding}, and legal NLP raises ethical limits of its own \citep{tsarapatsanis2021ethical}.

\section{Task and Data}
\label{sec:data}
The corpus holds 3{,}360 sponsored segments from child-facing YouTube channels \citep{bertaglia2026childsafeads}.
A segment is one sponsored passage inside a video, and each video contributes exactly one.
One segment together with its context fields is one datapoint, and the access levels of \S\ref{sec:levels-def} set how many of those fields a system may read.
Channels group the datapoints, a median of two each and up to 25.
Training holds 2{,}353 datapoints from 632 channels, development 504 datapoints from 154 channels, and test 503 from 153.
The splits are channel-disjoint, so a system never sees a test channel during training.
Development labels are public and test labels are withheld.
Two facts are given per datapoint and not re-assessed, that the channel is child-facing and that the segment is commercial.

\subsection{Subtasks and Labels}
\label{sec:subtasks}
The three subtasks label the same segment on three axes, the type of commercial relationship (ST1), what is being sold (ST2), and where the segment risks non-compliance (ST3).
\textbf{ST1} assigns one commercial type out of five, descended from the contract typology of the Consumer Rights Directive \citep{crd2011}.
\texttt{physical\_goods} ($47.0\%$) and \texttt{digital\_content\_or\_services} ($46.1\%$) cover $93\%$ of the labelled data, while \texttt{other} occurs twice in 2{,}857 datapoints.
\textbf{ST2} assigns product categories, 12 of them and $1.32$ per datapoint.
\textbf{ST3} assigns compliance risk flags, 8 of them and $1.35$ per datapoint, drawing mainly on the UCPD \citep{ucpd2005} and the AVMSD \citep{avmsd2010}.
Every label set is steeply skewed, so macro-$\fone$ depends heavily on a small number of rare-label predictions.
The ST3 taxonomy also carries structure a system must respect, since \texttt{no\_flag} and \texttt{insufficient\_context} each stand alone, and \texttt{undisclosed\_advertising} and \texttt{inadequate\_disclosure} are mutually exclusive, the distinction between disclosure absent and disclosure present but not recognisable.

\subsection{Access Levels}
\label{sec:levels-def}
The four access levels are cumulative.
L1 is the segment transcript (median ${\approx}300$ tokens).
L2 adds the video title, the description and the platform's paid-promotion label (${\approx}290$ tokens).
That label is the platform's declaration mechanism under AVMSD Article~28b(3)(c) \citep{avmsd2010}.
L3 adds the channel name (${\approx}7$ tokens).
L4 adds the text of the promoted product's resolved web page (${\approx}375$ tokens, with a heavy tail).
Every datapoint fits into 8{,}192 tokens without truncation at full access (maximum 8{,}039 tokens).

\section{System}
\label{sec:system}
Figure~\ref{fig:arch} shows the target shape, and Figure~\ref{fig:pipeline} in Appendix~\ref{app:pipeline} draws the same system end to end, from the input to the prediction.
This section builds it bottom-up, from backbones and class scopes through cross-validation to the nine-voter vote.

\subsection{Backbones and Adaptation Methods}
\label{sec:pool}
A configuration fixes the four axes of a voter, backbone, method, class scope and access level, and its name lists them in that order, with the subtask of an S or MCS, \texttt{backbone-method-scope-subtask-level}.
A \textit{voter} is one trained fold-model of a configuration (\S\ref{sec:cv}).
Table~\ref{tab:pool} names the four backbones and the seven methods.
The methods range from training the whole backbone to training nothing at all.
Across all four levels the pool holds 1{,}149 banked configurations, not all of them complete, with 5{,}612 finished fold-level prediction sets.

\begin{table}[h]
  \centering
  \fontsize{7.5}{9}\selectfont
  \setlength{\tabcolsep}{1pt}
  \begin{tabular*}{\columnwidth}{@{}ll@{\extracolsep{\fill}}r@{}}
    \toprule
    \multicolumn{3}{@{}l}{\textit{Backbones}} \\
    \midrule
    min    & Ministral-8B-Instruct-2410 \citep{mistral2024ministral} & {\color{black!55}\fontsize{6}{7}\selectfont dec} \\
    phi4   & Phi-4, 14B \citep{abdin2024phi4} & {\color{black!55}\fontsize{6}{7}\selectfont dec} \\
    ett1b  & ettin-encoder-1b \citep{weller2025ettin} & {\color{black!55}\fontsize{6}{7}\selectfont enc} \\
    eub610 & EuroBERT-610m \citep{boizard2025eurobert} & {\color{black!55}\fontsize{6}{7}\selectfont enc} \\
    \midrule
    \multicolumn{3}{@{}l}{\textit{Training methods}} \\
    \midrule
    SFT     & generative fine-tuning & {\color{black!55}\fontsize{6}{7}\selectfont dec} \\
    ClsHead & discriminative fine-tuning & {\color{black!55}\fontsize{6}{7}\selectfont dec} \\
    FT      & discriminative fine-tuning & {\color{black!55}\fontsize{6}{7}\selectfont enc} \\
    Base-LR/-ClsHead & light head on untrained backbones & {\color{black!55}\fontsize{6}{7}\selectfont dec/enc} \\
    SFTf-LR/-ClsHead & light head per subtask on the G & {\color{black!55}\fontsize{6}{7}\selectfont dec} \\
    FTf-LR/-ClsHead  & light head per subtask on the G & {\color{black!55}\fontsize{6}{7}\selectfont enc} \\
    OPRO    & optimised prompt \citep{yang2023opro} & {\color{black!55}\fontsize{6}{7}\selectfont dec} \\
    \bottomrule
  \end{tabular*}
  \caption{Backbones and training methods of the voter pool. dec and enc mark decoder and encoder.}
  \label{tab:pool}
\end{table}

\textbf{Trained backbones.}
SFT fine-tunes a decoder to generate the label as text.
ClsHead trains a classification head jointly with a decoder backbone, and FT does the same on an encoder.
The decoders train under 4-bit QLoRA \citep{dettmers2023qlora}, the encoders update all weights.

\textbf{Reused backbones.}
The frozen-backbone methods cache a backbone's embeddings once and fit a \textit{light head} on top, a logistic regression in the \texttt{-LR} variants and a small multi-layer perceptron (MLP) in the \texttt{-ClsHead} variants, which makes them the cheapest voters.
Both heads read the backbone's last hidden layer, taken from the final non-padding token on a decoder and mean-pooled over the tokens on an encoder.
The SFTf and FTf variants place such a head on a G backbone that SFT or FT has already tuned across all three subtasks (\S\ref{sec:scopes}), so a per-subtask head inherits that domain knowledge without a second backbone training.

\textbf{Untrained backbones.}
The Base variants fit the same heads on the untrained backbone, and OPRO optimises a prompt and trains nothing.

Only voters whose backbone had seen the task reached a deployed ensemble (Table~\ref{tab:submissions}), and the margin is wide.
OPRO peaks at an $\fcvtop$ of $0.61$ on ST1, $0.62$ on ST2 and $0.23$ on ST3, where the best trained method reaches $0.86$, $0.89$ and $0.69$.
The best head on an untrained backbone comes closer, yet still trails by a clear $0.15$, $0.07$ and $0.15$ on the three subtasks.
These methods therefore never reached a deployed ensemble, and eub610 never ranked high enough for one either.
Appendix~\ref{app:methods} compares every method at full access.

\subsection{Class Scopes}
\label{sec:scopes}
\begin{figure}[h]
\centering
\resizebox{\columnwidth}{!}{%
\begin{tikzpicture}[
  cell/.style={rectangle, rounded corners=1.5pt, line width=0.55pt,
    minimum width=1.7cm, minimum height=0.56cm, align=center,
    font=\fontsize{6.8}{8.2}\selectfont},
  chip/.style={rectangle, rounded corners=1.2pt, line width=0.45pt,
    minimum height=0.38cm, minimum width=1.55cm, align=center, inner sep=1.5pt,
    font=\fontsize{6.8}{8.2}\selectfont},
  rowlab/.style={anchor=east, align=right, font=\fontsize{7.5}{9}\selectfont\bfseries},
  rowsub/.style={anchor=east, align=right, font=\fontsize{6.3}{7.6}\selectfont\itshape, text=black!50},
  collab/.style={align=center, font=\fontsize{7.5}{9}\selectfont\bfseries, text=black!60},
  colsub/.style={align=center, font=\fontsize{6.3}{7.6}\selectfont\itshape, text=black!50},
  freq/.style={rectangle, draw, line width=0.4pt, minimum height=0.30cm,
    anchor=south west, inner sep=0pt},
  keepsub/.style={font=\fontsize{6.3}{7.6}\selectfont, text=clrSFT!45!black, align=center},
]
\def\cOne{-1.0}
\def\cTwo{0.8}
\def\cThree{2.6}
\node[collab] at (\cOne, 2.02) {ST1};
\node[colsub] at (\cOne, 1.62) {single-label\\5 classes};
\node[collab] at (\cTwo, 2.02) {ST2};
\node[colsub] at (\cTwo, 1.62) {multi-label\\12 labels};
\node[collab] at (\cThree, 2.02) {ST3};
\node[colsub] at (\cThree, 1.62) {multi-label\\8 flags};
\node[rowlab, text=clrCls!45!black] at (-2.0, 1.06) {Generalist (G)};
\node[rowsub] at (-2.0, 0.80) {one model, jointly};
\node[rectangle, rounded corners=2.5pt, draw=clrCls!60!black,
      dash pattern=on 1.2pt off 1.2pt,
      fill=clrCls!12, line width=0.7pt, minimum width=5.40cm,
      minimum height=0.60cm] at (\cTwo, 0.96) {};
\node[chip, draw=clrCls!55!black, fill=white] at (\cOne, 0.96) {all 5 labels};
\node[chip, draw=clrCls!55!black, fill=white] at (\cTwo, 0.96) {all 12 labels};
\node[chip, draw=clrCls!55!black, fill=white] at (\cThree, 0.96) {all 8 flags};
\node[rowlab, text=clrThird!40!black] at (-2.0, 0.08) {Specialist (S)};
\node[cell, dash pattern=on 1.2pt off 1.2pt, draw=clrThird!65!black, fill=clrThird!18] at (\cOne, 0.08) {all 5 labels};
\node[cell, dash pattern=on 1.2pt off 1.2pt, draw=clrThird!65!black, fill=clrThird!18] at (\cTwo, 0.08) {all 12 labels};
\node[cell, dash pattern=on 1.2pt off 1.2pt, draw=clrThird!65!black, fill=clrThird!18] at (\cThree, 0.08) {all 8 flags};
\node[rowlab, text=clrSFT!50!black] at (-2.0, -0.70) {MCS};
\node[rowsub] at (-2.0, -0.96) {min.-class specialist};
\foreach \c in {\cOne, \cTwo, \cThree}{
  \node[cell, dash pattern=on 1.2pt off 1.2pt, draw=clrSFT!62!black, fill=white] at (\c, -0.80) {};
}
\begin{scope}[shift={({\cOne-0.76}, -0.95)}]
  \foreach \x/\w in {0.000/0.707, 0.707/0.694}{
    \node[freq, draw=black!42, fill=black!8, minimum width=\w cm] at (\x, 0) {};
    \draw[black!32, line width=0.35pt] (\x, 0) -- ({\x+\w}, 0.30);
    \draw[black!32, line width=0.35pt] (\x, 0.30) -- ({\x+\w}, 0);
  }
  \foreach \x/\w in {1.401/0.077, 1.478/0.026, 1.504/0.016}{
    \node[freq, draw=clrSFT!60!black, fill=clrSFT!20, minimum width=\w cm] at (\x, 0) {};
  }
\end{scope}
\begin{scope}[shift={({\cTwo-0.76}, -0.95)}]
  \foreach \x/\w in {0.000/0.330, 0.330/0.253}{
    \node[freq, draw=black!42, fill=black!8, minimum width=\w cm] at (\x, 0) {};
    \draw[black!32, line width=0.35pt] (\x, 0) -- ({\x+\w}, 0.30);
    \draw[black!32, line width=0.35pt] (\x, 0.30) -- ({\x+\w}, 0);
  }
  \foreach \x/\w in {0.583/0.199, 0.782/0.136, 0.918/0.128, 1.046/0.127, 1.172/0.126, 1.298/0.062, 1.360/0.061, 1.421/0.044, 1.465/0.034, 1.499/0.016}{
    \node[freq, draw=clrSFT!60!black, fill=clrSFT!20, minimum width=\w cm] at (\x, 0) {};
  }
\end{scope}
\begin{scope}[shift={({\cThree-0.76}, -0.95)}]
  \foreach \x/\w in {0.000/0.597}{
    \node[freq, draw=black!42, fill=black!8, minimum width=\w cm] at (\x, 0) {};
    \draw[black!32, line width=0.35pt] (\x, 0) -- ({\x+\w}, 0.30);
    \draw[black!32, line width=0.35pt] (\x, 0.30) -- ({\x+\w}, 0);
  }
  \foreach \x/\w in {0.597/0.284, 0.881/0.255, 1.136/0.166, 1.301/0.148, 1.450/0.029, 1.479/0.017, 1.496/0.016}{
    \node[freq, draw=clrSFT!60!black, fill=clrSFT!20, minimum width=\w cm] at (\x, 0) {};
  }
\end{scope}
\node[keepsub] at (\cOne, -1.30) {3 of 5 kept};
\node[keepsub] at (\cTwo, -1.30) {10 of 12 kept};
\node[keepsub] at (\cThree, -1.30) {7 of 8 kept};
\node[rectangle, rounded corners=1.5pt, draw=black!55,
      dash pattern=on 1.2pt off 1.2pt, line width=0.5pt,
      minimum width=0.55cm, minimum height=0.28cm] at (-4.05, 2.02) {};
\node[anchor=west, font=\fontsize{6.3}{7.6}\selectfont, text=black!60]
  at (-3.73, 2.02) {= one trained model};
\end{tikzpicture}%
}
\caption{What each scope trains on. Crossed blocks are dropped majority labels, and block widths are proportional to label frequency.}
\label{fig:scopes}
\end{figure}
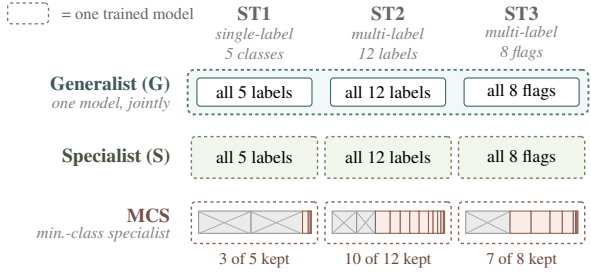

The class scopes are the axis the architecture is built around.
A \textbf{generalist} (G) trains jointly on all three subtasks, and the prompt selects the subtask at inference (Appendix~\ref{app:prompts}).
A \textbf{specialist} (S) trains on one subtask.
A \textbf{minority-class specialist} (MCS) trains on one subtask with the most frequent labels removed, so G and S are the \textit{full-label} scopes.
Starting from the most frequent label, labels are added up until their combined share exceeds half the label mass, and exactly these labels are dropped.
That rule drops \texttt{physical\_goods} and \texttt{digital\_content\_or\_services} on ST1, $93\%$ of the mass between them.
On ST2 it drops \texttt{apps} and \texttt{hardware\_electronics} ($51\%$), and on ST3 \texttt{misleading\_claim} ($54\%$).
What removal costs differs by task.
On the single-label ST1 the datapoints carrying those classes leave the training set with them.
On the multi-label ST2 and ST3 only the label columns go, so majority-only datapoints stay as all-negative training signal against over-firing.
An MCS is scored only on its own labels, so its score sits on a different scale and never compares against G or S.
Figure~\ref{fig:scopes} shows which subtasks and labels each scope trains on.

\subsection{Cross-Validation and Branches}
\label{sec:cv}
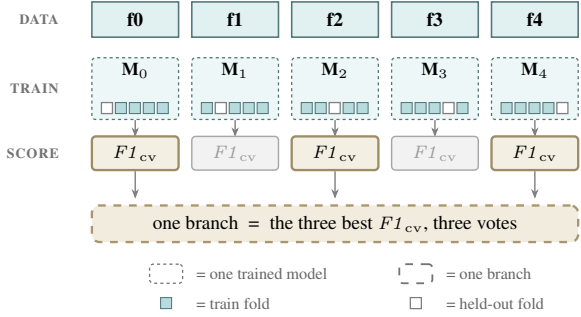
\begin{figure}[h]
\centering
\resizebox{\columnwidth}{!}{%
\begin{tikzpicture}[
  fold/.style={rectangle, draw=clrCls!60!black, fill=clrCls!25, line width=0.6pt},
  vbox/.style={rectangle, draw=clrCls!60!black, fill=clrCls!14, rounded corners=1.5pt,
    dash pattern=on 1.2pt off 1.2pt,
    line width=0.5pt, minimum width=1.18cm, minimum height=0.84cm},
  fchip/.style={rectangle, rounded corners=1.5pt, line width=0.55pt,
    minimum width=1.18cm, minimum height=0.42cm, align=center,
    font=\fontsize{7}{8.4}\selectfont},
  hdr/.style={font=\fontsize{6.8}{8.2}\selectfont\bfseries, text=black!55},
  edge/.style={-{Stealth[length=3.8pt]}, line width=0.55pt, black!50},
  major/.style={rectangle, rounded corners=3pt, line width=0.9pt,
    align=center, font=\fontsize{7.5}{9}\selectfont, inner ysep=4pt, inner xsep=8pt},
  leg/.style={anchor=west, font=\fontsize{6.3}{7.6}\selectfont, text=black!60},
]
\foreach \i in {0,...,4}{
  \draw[fold] ({\i*1.36 - 3.31}, 2.80) rectangle ++(1.18, 0.48);
  \node[font=\fontsize{7.5}{9}\selectfont\bfseries] at ({\i*1.36 - 2.72}, 3.04) {f\i};
}
\node[hdr, anchor=east] at (-3.62, 3.04) {\textsc{data}};
\node[hdr, anchor=east] at (-3.62, 2.10) {\textsc{train}};
\foreach \i in {0,...,4}{
  \node[vbox] (m\i) at ({\i*1.36 - 2.72}, 2.10) {};
  \node[font=\fontsize{7}{8.4}\selectfont\bfseries, anchor=north, inner sep=2pt]
    at (m\i.north) {M$_\i$};
  \foreach \j in {0,...,4}{
    \pgfmathsetmacro{\cx}{\i*1.36 - 2.72 - 0.46 + \j*0.19}
    \ifnum\j=\i\relax
      \draw[black!50, fill=white, line width=0.5pt] (\cx, 1.76) rectangle ++(0.15, 0.15);
    \else
      \draw[clrCls!60!black, fill=clrCls!60, line width=0.4pt]
        (\cx, 1.76) rectangle ++(0.15, 0.15);
    \fi
  }
}
\node[hdr, anchor=east] at (-3.62, 1.22) {\textsc{score}};
\foreach \i/\keep in {0/1, 1/0, 2/1, 3/0, 4/1}{
  \draw[edge] (m\i.south) -- ({\i*1.36 - 2.72}, 1.44);
  \ifnum\keep=1\relax
    \node[fchip, draw=clrEns!75!black, fill=clrEns!24, line width=0.9pt]
      (s\i) at ({\i*1.36 - 2.72}, 1.22) {$\fcv$};
  \else
    \node[fchip, draw=black!35, fill=black!5, text=black!40]
      (s\i) at ({\i*1.36 - 2.72}, 1.22) {$\fcv$};
  \fi
}
\foreach \i in {0, 2, 4}{ \draw[edge] (s\i.south) -- ({\i*1.36 - 2.72}, 0.60); }
\node[major, dashed, draw=clrEns!80!black, fill=clrEns!28, minimum width=6.62cm]
  at (0, 0.26) {one branch \;=\; the three best $\fcv$, three votes};
\node[rectangle, rounded corners=1.5pt, draw=black!55,
      dash pattern=on 1.2pt off 1.2pt, line width=0.5pt,
      minimum width=0.45cm, minimum height=0.26cm] at (-2.30, -0.42) {};
\node[leg] at (-2.02, -0.42) {= one trained model};
\node[rectangle, rounded corners=1.5pt, draw=black!55, dashed, line width=0.7pt,
      minimum width=0.45cm, minimum height=0.26cm] at (1.10, -0.42) {};
\node[leg] at (1.38, -0.42) {= one branch};
\draw[clrCls!60!black, fill=clrCls!60, line width=0.4pt]
  (-2.375, -0.895) rectangle ++(0.15, 0.15);
\node[leg] at (-2.02, -0.82) {= train fold};
\draw[black!50, fill=white, line width=0.5pt]
  (1.025, -0.895) rectangle ++(0.15, 0.15);
\node[leg] at (1.38, -0.82) {= held-out fold};
\end{tikzpicture}%
}
\caption{A CV5 configuration. Each held-out fold scores its model. The three best $\fcv$ form a branch.}
\label{fig:cv5}
\end{figure}

Every configuration trains five times under five-fold cross-validation (CV5).
The folds are grouped by channel, since one channel's segments share vocabulary, product mix and disclosure habits, and a random split would leak what the channel-disjoint test set withholds.
Run $i$ trains on four folds and is scored on the held-out fold $i$, which gives the fold-model's $\fcv$.
The mean of the three best folds, $\fcvtop$, is the configuration's selection signal (Figure~\ref{fig:cv5}).

A \textbf{branch} deploys the three fold-models with the best held-out $\fcv$, the same three that form its $\fcvtop$, and leaves two folds out.
We deploy three rather than five mainly because nine voters cost less to run than fifteen.
We also hope that leaving out different fold-models in different branches adds variance to the ensemble, since the held-out fold a branch drops can be the one another branch keeps.
Three branches choosing independently could leave the same fold out, so we require the three branches together to cover all five folds, and a branch then keeps its best two folds, adds the missing one and takes its $\fcvtop$ over those three.
Appendix~\ref{app:training} gives the training hyperparameters and the threshold tuning.

\begin{table*}[t!]
  \centering
  \footnotesize
  \renewcommand{\arraystretch}{0.94}
  \setlength{\tabcolsep}{3pt}
  \begin{tabular*}{\textwidth}{@{\extracolsep{\fill}}l >{\raggedright\arraybackslash}p{11.6cm} c r@{}}
    \toprule
    & \textbf{Voter composition} & mean $\boldsymbol{\fcvtop}$ & $\boldsymbol{\ftest}$ \\
    \midrule
    \multicolumn{4}{@{}l}{\textbf{Submission 1}} \\
    ST1 & phi4-SFT-G + ett1b-FT-S + phi4-SFTf-LR-MCS-L1 & .7877 & .5944 \\
    ST2 & phi4-SFT-G + ett1b-FT-S + phi4-SFTf-LR-MCS    & .8624 & .8034 \\
    ST3 & phi4-SFT-G + ett1b-FT-S + phi4-SFTf-LR-MCS    & .6427 & .5954 \\
    \midrule
    \multicolumn{4}{@{}l}{\textbf{Submission 2}} \\
    ST1 & phi4-SFT-S + 2$\times$\,min-ClsHead-S & .8232 & .6205 \\
    ST2 & min-ClsHead-S + phi4-ClsHead-S + phi4-SFT-S & .8715 & .8204 \\
    ST3 & phi4-SFTf-LR-S-L123 + phi4-SFT-G + ett1b-FTf-ClsHead-S-L12 & .6656 & .6512 \\
    \midrule
    \multicolumn{4}{@{}l}{\textbf{Submission 3}} \\
    ST1 & min-ClsHead-S-L12 + phi4-ClsHead-G + phi4-SFT-S & .7926 & .5339 \\
    ST2 & min-ClsHead-S + phi4-ClsHead-S + phi4-SFT-S + phi4-ClsHead-MCS & .8378 & \textbf{.8243} \\
    ST3 & phi4-SFTf-LR-S-L123 + phi4-SFT-G + ett1b-FTf-ClsHead-S-L12 & .6656 & .6483 \\
    \midrule
    \multicolumn{4}{@{}l}{\textbf{Submission 4}} \\
    ST1 & phi4-SFTf-LR-G + phi4-SFT-S + min-ClsHead-S & .7922 & \textbf{.6464} \\
    ST2 & min-ClsHead-G + ett1b-FTf-ClsHead-S + min-SFT-S & .8458 & .7719 \\
    ST3 & phi4-SFT-G + phi4-SFTf-LR-S + ett1b-FTf-ClsHead-S & .6688 & \textbf{.6530} \\
    \bottomrule
  \end{tabular*}
  \caption{Voter compositions of Submissions 1 to 4, mean $\fcvtop$ over the deployed branches and hidden-test macro-$\fone$. Bold marks the best test score per subtask. 2$\times$ marks two separately trained configurations of the same recipe. Where an ensemble holds an MCS branch, the mean includes its own-scale score and is descriptive only.}
  \label{tab:submissions}
\end{table*}

\subsection{Vote Aggregation}
\label{sec:voting}
An \textbf{ensemble} comprises three branches, each casting three of the nine votes.
Figure~\ref{fig:arch} states the aggregation formally.
Every voter returns a set of labels, exactly one class on the single-label ST1 and possibly several on ST2 and ST3.
For each label the ensemble counts the voters that named it and keeps the labels reaching five of nine, so a voter that does not name a label counts against it.
The task-specific rules of \S\ref{sec:rules} then turn that set into the prediction.
On ST1 each voter's single vote goes to one class, so the classes compete and the nine can scatter until none reaches five.
On ST2 and ST3 a voter can name several labels, so labels do not compete and five of nine acts as an independent yes/no threshold per label.

An ensemble must also be structurally dissimilar, spanning at least two backbones and both prediction paradigms, generative and discriminative.
This rule is a hard constraint on the selection.
The intended template combines a G, an S and an MCS branch, and Submission 1 deployed it as such.
Because full-label branches ranked more reliably on every subtask, the later submissions combine G and S branches only, and an MCS reappears once, as an added fourth branch on ST2 (Table~\ref{tab:submissions}).
An MCS never names a majority label, and with the threshold fixed at five of nine its abstention counts as three votes against, so it can only arbitrate where the two full-label branches disagree.

\subsection{Taxonomy Rules}
\label{sec:rules}
In the vote formula of Figure~\ref{fig:arch}, $P_t$ applies the taxonomy rules of subtask $t$.
Its input is the raw vote result, the set of labels that reached five of nine votes.
The nine votes can split so that no label reaches five, leaving the set empty, and on ST3 the set can contradict the taxonomy.
Each subtask repairs this with its own rules.
ST1 requires exactly one class.
$P_1$ returns the class that reached five votes, and when no class did, it falls back to the most frequent class, \texttt{physical\_goods}.
A 4--3--2 split therefore yields \texttt{physical\_goods} rather than the leading class.
All submissions except Submission 3 suppress \texttt{other} ($0.07\%$ of the labelled data).
Neither $P_2$ nor $P_3$ ever returns an empty set, because when no label reaches five, the label with the largest vote count is taken.
ST2 has no taxonomy constraints, so $P_2$ changes nothing else.
On ST3 the voters decide every flag independently, so the vote result can be contradictory.
$P_3$ enforces the taxonomy of \S\ref{sec:subtasks}, where \texttt{no\_flag} and \texttt{insufficient\_context} stand alone, and the two disclosure flags exclude each other.
The official validity checker does not test these constraints.

\section{Submissions and Results}
\label{sec:results}
All five allowed uploads were used, each declaring access level L1234.

\subsection{Submission Strategy}
\label{sec:uploads}
Table~\ref{tab:submissions} lists the voter compositions of Submissions 1--4 against their hidden-test scores in the nomenclature of \S\ref{sec:pool}, with subtask tags omitted and L1234 assumed unless marked.

\textbf{Submission 1} is the plain G\,$\times$\,S\,$\times$\,MCS template, phi4-SFT-G, ett1b-FT-S and phi4-SFTf-LR-MCS per subtask ($\fcvtop$ $0.7643$).
\textbf{Submission 2} selects each ensemble by top $\fcvtop$ under the dissimilarity rule of \S\ref{sec:voting} ($\fcvtop$ mean $0.7868$).
\textbf{Submission 3} probes whether lifting the minority labels pays off.
ST1 moves to a development-supported ensemble in which a minority class wins from three of nine votes and \texttt{other} from two, ST2 adds phi4-ClsHead-MCS as a fourth branch, twelve voters, where a minority label also fires when the three MCS votes and at least four of the nine base votes agree, and ST3 keeps its voters and merely lowers the firing threshold to four of nine votes.
\textbf{Submission 4} refits the selected configurations in channel-disjoint CV5 over the combined training and development pool (refit $\fcvtop$ mean $0.7690$).
For the fifth and final upload we combined the best already scored fields per subtask unchanged, ST1 and ST3 from Submission 4 and ST2 from Submission 3, which scored $0.7079$ and became our leaderboard entry.
The ensembles of all five uploads were hand-picked from the branches with the highest $\fcvtop$, taking those that are at the same time as different as possible in backbone, method and scope (\S\ref{sec:voting}), with the one development-set exception of \S\ref{sec:transfer}.
Each upload after the first was composed with the test scores of the earlier ones known.

\subsection{Leaderboard and the ST1 Gap}
\label{sec:gap}
Our entry holds the best ST2, ST3 and ST3-family scores of all 22 final entries, and on the official mean it places third at $0.7079$.
On the won subtasks the entry leads the best other entry by $+0.021$ (ST2) and $+0.038$ (ST3).
On ST1 the system falls $0.203$ short of the best leaderboard entry.
We have not established why the ensemble that wins both multi-label subtasks falls short on the single-label one, and examining this is future work.

\subsection{Validation on the Hidden Test}
\label{sec:transfer}
\textbf{Selection.}
The $\fcvtop$ signal carried the deployed choices, with one exception described below, and one measurement suggests why the development set was less suited to this role.
Splitting the development set's 504 datapoints into two channel-disjoint halves, we scored the same candidate ensembles on both.
Our hypothesis was that, if the development set measured anything stable, the candidates leading on one half would also lead on the other.
Instead the two rankings correlate at only $r{=}0.06$ (ST1), $0.03$ (ST2) and $0.37$ (ST3) across the 300 candidates, so which candidate looks best depends largely on the particular channels scoring it.
Rescored on the other half, the best candidate of one half loses $0.267$ on ST1.
A development set this small cannot predict the test ranking, and we read it as a transfer check only.
The hidden test confirmed this.
Submission 3 chose its ST1 ensemble by development score, and it scored best on the development set and worst on test ($0.5339$).
The cross-validation pick fared better and ranked ST2 correctly for Submission 2.
The field-best ST2 score of $0.8243$ comes from a twelve-voter variant we tried once in Submission 3, a fourth MCS branch added to the nine, and its gain of $0.004$ over the nine-voter $0.8204$ is within noise, so the nine-voter ensemble stays our reference system.

\textbf{The deployed changes.}
Each rests on one hidden-test observation, so the deltas below are indicative rather than established.
The five-vote threshold decides how many labels a datapoint receives, and the deployed systems average $1.32$ labels on ST2 and $1.33$ on ST3, close to the $1.32$ and $1.35$ of the labelled data, which we read as one indication that the threshold is placed correctly.
Lowering it to four of nine looked right on the development set, where ST3 rose from $0.7151$ to $0.7322$ while ST2 would have fallen from $0.7785$ to $0.7688$, so Submission 3 lowered it for ST3 only.
The test set reversed that gain, and Submission 3's four-vote ST3 scored $0.6483$ against Submission 2's $0.6512$.
The refit over training and development split the same way, helping ST1 by $0.026$ and ST3 by $0.002$ while costing $0.049$ on ST2, so refitting and composition are not separable choices.
Plurality voting would be the more natural rule for ST1, because a single-label task needs one winning class and the fallback shifts every weak winner towards the dominant class.
Replaying the final votes changes three of 503 predictions, all unique 4--3--2 splits, and the withheld test labels leave those three unscorable.

\section{What Each Access Level Contributes}
\label{sec:levels}
Table~\ref{tab:pooled-levels} covers every G and every full-label S configuration trained on the complete training pool.
It groups them by exact access level and reports the pooled mean and 95\% confidence interval of the branch-selection score $\fcvtop$.
A G contributes one observation per subtask and an S one.
MCS configurations are excluded because their reduced-label score is on a different scale (\S\ref{sec:scopes}).
Task-specific values are in Appendix~\ref{app:level-table}.

\begin{table}[h]
  \centering
  \small
  \setlength{\tabcolsep}{4pt}
  \begin{tabular*}{\columnwidth}{@{\extracolsep{\fill}}lccc@{}}
    \toprule
    \textbf{Access} & \textbf{Mean $\fcvtop$} & \textbf{95\% t-CI} & $\boldsymbol{n}$ \\
    \midrule
    L1    & $0.582$ & [$0.563$, $0.600$] & 148 \\
    L12   & $0.614$ & [$0.592$, $0.637$] & 156 \\
    L123  & $0.609$ & [$0.587$, $0.631$] & 157 \\
    L1234 & $0.681$ & [$0.657$, $0.705$] & 157 \\
    \bottomrule
  \end{tabular*}
  \caption{Pooled voter-level $\fcvtop$ by access level with 95\% t-CIs.}
  \label{tab:pooled-levels}
\end{table}

Video context at L2 raises the pooled mean by $0.033$, the channel name at L3 changes it by only $-0.006$, and product-page text at L4 yields the largest increase, $+0.072$.
The sequence is therefore not monotone, and the L3 step is indistinguishable from zero in all three of our measurements.
The L4 increase is concentrated in ST1 ($+0.080$) and ST2 ($+0.116$), and ST3 rises by only $0.013$.
The intervals overlap across the successive L1--L3 comparisons but not between L123 and L1234, and given the heterogeneous configurations we treat this as descriptive rather than causal evidence.
A stricter comparison over 108 matched configuration--subtask series reproduces the ordering ($+0.026$, $-0.010$, $+0.073$).

On the deployed nine-voter system, the same three steps give $+0.074$ $[{+}0.024, {+}0.106]$ for L2, $+0.005$ $[{-}0.011, {+}0.019]$ for L3 and $+0.031$ $[{+}0.013, {+}0.050]$ for L4 on the development set (fifty best ensembles per level, paired bootstrap over the 154 channels, 2{,}000 draws).
L2 and L4 change places, because nine votes at L12 already recover much of what a single L1234 voter holds alone.

\section{Cost at Test-Set Scale}
\label{sec:cost}
Table~\ref{tab:cost} prices Submission~2, our strongest complete nine-voter system and reference implementation.
We exclude the twelve-voter ST2 extension, whose hidden-test gain was only $0.0039$ (\S\ref{sec:transfer}).
Training covers all five CV folds of the three retained configurations per subtask, for frozen-backbone voters including the backbone training they reuse.
Inference deploys the selected three folds of each, nine voters in total.

\begin{table}[h]
  \centering
  \small
  \setlength{\tabcolsep}{2.5pt}
  \begin{tabular*}{\columnwidth}{@{\extracolsep{\fill}}lccc@{}}
    \toprule
    \textbf{Subtask} & \shortstack{\textbf{Train}\\\textbf{GPU-h}} & \shortstack{\textbf{Test}\\\textbf{s/segment}} & \shortstack{\textbf{Test}\\\textbf{GPU-h (503)}} \\
    \midrule
    ST1 & $\approx170$ & 5.4 & 0.75 \\
    ST2 & $\approx153$ & 5.8 & 0.81 \\
    ST3 & $\approx251$ & 9.6 & 1.34 \\
    \midrule
    full system & $\approx574$ & 20.8 & 2.91 \\
    \bottomrule
  \end{tabular*}
  \caption{Training and test-scale inference cost of Submission~2.}
  \label{tab:cost}
\end{table}

The training column sums logged one-GPU wall time for the fold-models behind each subtask, excluding the CPU-only head fits.
The jobs ran on mixed A100, H200 and L40S accelerators, so these are raw GPU-hours.
The inference columns are reconstructed on one A100-80GB from measured single-pass wall times and the deployed folds, with ${\approx}2.02$\,s per segment for frozen Phi-4, $0.25$\,s for frozen ettin-1b and $0.52$\,s for a Ministral-8B pass.
No step uses a paid API.
The input prompt is subtask-specific, so only model loading amortises across subtasks.

\section{Conclusion}
\label{sec:conclusion}
Nine structurally dissimilar voters per subtask, selected on channel-disjoint cross-validation, win two of the three subtasks with the field-best ST2 and ST3 scores and place third on the mean.
The system is weak on the single-label ST1, which decided the mean ranking, and examining why is future work.
For the configurations we trained on this corpus, the access comparison reads as follows.
Video context (L2) raised the pooled voter score on every subtask.
With our system we could not measure any gain from the channel name (L3).
The product page (L4) gave the voters the largest gain, concentrated in commercial type (ST1) and product category (ST2).
For the compliance flags (ST3) our system gained little beyond L2.
These are descriptive findings for the evaluated systems, not causal estimates (\S\ref{sec:levels}).
At full access the ensemble labels the 503-datapoint test set in under three GPU-hours on one A100.
The third result concerns model selection.
We do not claim that cross-validation is a demonstrably better selector than the development set.
It was the pragmatic choice, because it scores every candidate on all 2{,}353 training datapoints through their held-out folds, whereas the development set offers only 504.
The ensembles it selected for ST2 and ST3 turned out to be the best in the field, but its best ST1 pick remained weak, so a good selection signal did not make up for a weak candidate pool on ST1.
The 504-datapoint development set could not separate the top candidates.

\section*{Limitations}
The access-level comparison in Table~\ref{tab:pooled-levels} aggregates heterogeneous configurations and is descriptive rather than a controlled ablation.
Its t-confidence intervals assume independent observations.
Shared folds and related configurations break that assumption, and the matched comparison is a robustness check, not an independent test set.
Internal $\fcvtop$ values are optimistic in level, since thresholds are tuned on the folds that score them.
They are the system's branch-selection signal rather than an estimate of hidden-test performance.
No submission went to a single-voter baseline, so the ensemble's gain over its best member is unmeasured on test.
We also did not measure the ensemble's gain over its best member on cross-validation, did not test whether plurality voting improves ST1 in validation, and did not compare how stably cross-validation and the development set rank candidates.
The flags also encode judgments the text pipeline cannot observe.
Disclosure adequacy turns on whether a child recognises the segment as advertising, which depends on visual presentation and timing absent from transcripts.
The low split-half reliabilities bound what 504 development datapoints can certify, and our scores measure agreement with one operationalisation of the law, not compliance.
Each submission is a single hidden-test observation, so per-change attributions such as the MCS cast's $+0.004$ and the refit deltas are indicative, not established.
The test labels are withheld, so we can neither verify Submission 3's two \texttt{other} predictions nor separate that override from the simultaneous ensemble switch.
All results are for English-language YouTube content and this flag taxonomy.

\section*{Ethics Statement}
The dataset derives from SponsorBlock (CC BY-NC-SA 4.0) and is used only for this shared task.
We do not re-identify or contact creators, we report no result that marks an individual creator as non-compliant, and all figures are aggregate.
The compliance labels are a research benchmark, not legal advice and not a finding of unlawfulness.
A system whose ST1 score can lose a fifth of its value to one rare class should inform human review, not replace it \citep{steigerwald2025caia,steigerwald2026ethics}.
No test datapoint was manually labelled.

\section*{Acknowledgments}
Funded by the Deutsche Forschungsgemeinschaft (DFG, German Research Foundation), FIP 160, Project-ID 549142762.
Most training and inference ran on the GPU cluster of the Center for Artificial Intelligence (KIZ) of Technische Hochschule N\"urnberg.
The authors gratefully acknowledge the scientific support and HPC resources provided by the Erlangen National High Performance Computing Center (NHR@FAU) of the Friedrich-Alexander-Universit\"at Erlangen-N\"urnberg (FAU) under the BayernKI project v148eb.
BayernKI funding is provided by Bavarian state authorities.

\bibliography{custom}

\appendix
\begin{figure*}[t]
\centering
\resizebox{\textwidth}{!}{%
\begin{tikzpicture}[
  x=1cm, y=1cm,
  panel/.style={rectangle, rounded corners=3.5pt, draw=black!20, fill=black!2,
    line width=0.6pt},
  ptitle/.style={anchor=west, font=\fontsize{7}{8.4}\selectfont\bfseries,
    text=black!72},
  pnum/.style={circle, draw=black!35, fill=white, line width=0.5pt,
    inner sep=0.6pt, minimum size=0.32cm,
    font=\fontsize{6}{7.2}\selectfont\bfseries, text=black!60},
  box/.style={rectangle, rounded corners=2pt, line width=0.6pt, align=center,
    inner sep=2pt, font=\fontsize{6.3}{7.6}\selectfont},
  chip/.style={rectangle, rounded corners=1.5pt, line width=0.5pt, align=center,
    inner sep=1.4pt, font=\fontsize{6}{7.2}\selectfont},
  ann/.style={font=\fontsize{5.8}{7}\selectfont, text=black!58, align=center},
  annl/.style={font=\fontsize{5.8}{7}\selectfont, text=black!58, align=left,
    anchor=west},
  axlab/.style={font=\fontsize{6}{7.2}\selectfont\bfseries, text=black!50,
    anchor=west},
  rlab/.style={font=\fontsize{6.3}{7.6}\selectfont\bfseries, anchor=east},
  edge/.style={-{Stealth[length=4.5pt]}, line width=0.8pt, black!45},
  fine/.style={-{Stealth[length=3pt]}, line width=0.5pt, black!45},
]
\node[panel, fit={(0,2.55) (3.60,8.95)}] (pA) {};
\node[panel, fit={(3.90,2.55) (8.30,8.95)}] (pB) {};
\node[panel, fit={(8.60,2.55) (11.95,8.95)}] (pC) {};
\node[panel, fit={(12.25,2.55) (15.80,8.95)}] (pD) {};
\node[pnum] at (0.24,8.71) {a};  \node[ptitle] at (0.46,8.71) {Input and access level};
\node[pnum] at (4.14,8.71) {b};  \node[ptitle] at (4.36,8.71) {Voter pool};
\node[pnum] at (8.84,8.71) {c};  \node[ptitle] at (9.06,8.71) {Cross-validation};
\node[pnum] at (12.49,8.71) {d}; \node[ptitle] at (12.71,8.71) {Nine voters};
\foreach \xl/\xr in {0.10/3.50, 4.00/8.20, 8.70/11.85, 12.35/15.70}{
  \draw[black!14, line width=0.5pt] (\xl,8.48) -- (\xr,8.48);
}
\draw[edge] (3.60,5.55) -- (3.90,5.55);
\draw[edge] (8.30,5.55) -- (8.60,5.55);
\draw[edge] (11.95,5.55) -- (12.25,5.55);
\node[box, draw=black!45, fill=white, minimum width=3.20cm, minimum height=0.56cm]
  at (1.80,8.06) {sponsored segment};
\draw[fine] (1.80,7.71) -- (1.80,7.51);
\foreach \lv/\y in {L1/7.28, L12/6.86, L123/6.44, L1234/6.02}{
  \node[rlab, font=\fontsize{5.4}{6.5}\selectfont\bfseries, text=black!75] at (0.68,\y) {\lv};
}
\foreach \y/\n in {7.28/1, 6.86/2, 6.44/3, 6.02/4}{
  \fill[clrCls!55] (0.76,{\y-0.125}) rectangle ({0.76+0.802},{\y+0.125});
  \ifnum\n>1 \fill[clrThird!62] ({0.76+0.802},{\y-0.125}) rectangle ({0.76+1.577},{\y+0.125});\fi
  \ifnum\n>2 \fill[clrEns!85] ({0.76+1.577},{\y-0.125}) rectangle ({0.76+1.596},{\y+0.125});\fi
  \ifnum\n>3 \fill[clrSFT!62] ({0.76+1.596},{\y-0.125}) rectangle ({0.76+2.599},{\y+0.125});\fi
  \draw[black!30, line width=0.35pt] (0.76,{\y-0.125}) rectangle (3.36,{\y+0.125});
}
\node[ann, anchor=east, text=black!42] at (3.36,5.76) {cumulative, median tokens};
\foreach \cy/\clr/\txt in {5.46/clrCls/{transcript $\approx$300},
                           5.18/clrThird/{$+$ title, description,},
                           4.62/clrEns/{$+$ channel name $\approx$7},
                           4.34/clrSFT/{$+$ product page text $\approx$375}}{
  \fill[\clr!62] (0.28,{\cy-0.065}) rectangle (0.46,{\cy+0.065});
  \draw[black!30, line width=0.3pt] (0.28,{\cy-0.065}) rectangle (0.46,{\cy+0.065});
  \node[annl] at (0.52,\cy) {\txt};
}
\node[annl] at (0.52,4.92) {\hphantom{$+$ }paid-promotion label $\approx$290};
\draw[fine] (1.80,4.12) -- (1.80,3.92);
\node[box, draw=black!45, fill=white, minimum width=3.20cm, minimum height=0.60cm]
  at (1.80,3.60) {prompt builder, one level cap};
\draw[fine] (1.80,3.30) -- (1.80,3.10);
\node[chip, draw=black!40, fill=black!4, minimum width=2.3cm] at (1.80,2.90)
  {prompt $\leq 8{,}192$ tokens};
\node[axlab] at (4.05,8.26) {backbone};
\foreach \x/\t in {4.42/min, 5.36/phi4, 6.30/ett1b, 7.24/eub610}{
  \node[chip, draw=black!45, fill=white, minimum width=0.84cm] at (\x,7.94) {\t};
}
\node[ann, text=black!42] at (4.89,7.66) {decoder};
\node[ann, text=black!42] at (6.77,7.66) {encoder};
\node[axlab] at (4.05,7.36) {training method};
\node[chip, draw=black!45, fill=white, anchor=west,
  font=\fontsize{5}{6}\selectfont] (mA) at (4.05,7.04) {SFT $\cdot$ ClsHead $\cdot$ FT};
\node[chip, draw=black!45, fill=white, right=0.07cm of mA,
  font=\fontsize{5}{6}\selectfont] (mB) {SFTf-/FTf- heads};
\node[chip, draw=black!35, fill=black!5, right=0.07cm of mB,
  font=\fontsize{5}{6}\selectfont] (mC) {Base $\cdot$ OPRO};
\node[ann, text=black!42, font=\fontsize{5.3}{6.4}\selectfont] at (mA.south) [below=0.5pt] {trained backbone};
\node[ann, text=black!42, font=\fontsize{5.3}{6.4}\selectfont] at (mB.south) [below=0.5pt] {frozen backbone};
\node[ann, text=black!42, font=\fontsize{5.3}{6.4}\selectfont] at (mC.south) [below=0.5pt] {untrained};
\node[axlab] at (4.05,6.44) {class scope};
\foreach \x/\st in {5.10/ST1, 6.30/ST2, 7.50/ST3}{
  \node[ann, text=black!58] at (\x,6.16) {\textbf{\st}};
}
\node[rlab, text=clrCls!45!black, font=\fontsize{5.4}{6.5}\selectfont\bfseries] at (4.50,5.86) {G};
\node[rectangle, rounded corners=2pt, draw=clrCls!60!black, fill=clrCls!14,
  dash pattern=on 1.2pt off 1.2pt, line width=0.6pt,
  minimum width=3.55cm, minimum height=0.34cm] at (6.30,5.86) {};
\foreach \x/\t in {5.10/{all 5}, 6.30/{all 12}, 7.50/{all 8}}{
  \node[chip, draw=clrCls!55!black, fill=white, minimum width=1.02cm] at (\x,5.86) {\t};
}
\node[rlab, text=clrThird!40!black, font=\fontsize{5.4}{6.5}\selectfont\bfseries] at (4.50,5.26) {S};
\foreach \x/\t in {5.10/{all 5}, 6.30/{all 12}, 7.50/{all 8}}{
  \node[chip, draw=clrThird!62!black, fill=clrThird!18,
    dash pattern=on 1.2pt off 1.2pt, minimum width=1.02cm] at (\x,5.26) {\t};
}
\node[rlab, text=clrSFT!50!black, font=\fontsize{5.4}{6.5}\selectfont\bfseries] at (4.50,4.66) {MCS};
\foreach \x/\ww/\t in {5.10/0.949/{3 of 5}, 6.30/0.520/{10 of 12}, 7.50/0.551/{7 of 8}}{
  \fill[black!8] ({\x-0.51},4.51) rectangle ({\x-0.51+\ww},4.81);
  \draw[black!32, line width=0.3pt] ({\x-0.51},4.51) -- ({\x-0.51+\ww},4.81);
  \draw[black!32, line width=0.3pt] ({\x-0.51},4.81) -- ({\x-0.51+\ww},4.51);
  \fill[clrSFT!22] ({\x-0.51+\ww},4.51) rectangle ({\x+0.51},4.81);
  \draw[clrSFT!62!black, dash pattern=on 1.2pt off 1.2pt, line width=0.5pt]
    ({\x-0.51},4.51) rectangle ({\x+0.51},4.81);
  \node[ann, text=clrSFT!45!black] at (\x,4.34) {\t};
}
\draw[black!15, line width=0.5pt] (4.10,3.86) -- (8.10,3.86);
\node[annl, font=\fontsize{5.4}{6.5}\selectfont] at (4.10,3.66) {4 backbones, 7 methods, 3 scopes, 4 access levels};
\node[annl, font=\fontsize{5.4}{6.5}\selectfont] at (4.10,3.44) {$\rightarrow$ \textbf{1{,}149} configurations, 5{,}612 fold-level sets};
\node[chip, draw=black!40, fill=black!4, minimum width=3.0cm] at (6.10,2.98)
  {\texttt{backbone-method-scope-subtask-level}\\[-1pt]{\fontsize{5.2}{6.2}\selectfont$=$ one voter recipe}};
\node[ann] at (10.27,8.22) {training pool, five channel-disjoint folds};
\foreach \i in {0,...,4}{
  \draw[draw=clrCls!60!black, fill=clrCls!25, line width=0.5pt]
    ({9.13+\i*0.50},7.84) rectangle ++(0.42,0.26);
  \node[font=\fontsize{5.2}{6.2}\selectfont] at ({9.34+\i*0.50},7.97) {f\i};
}
\foreach \i/\y/\keep in {0/7.20/1, 1/6.76/0, 2/6.32/1, 3/5.88/0, 4/5.44/1}{
  \node[rlab, font=\fontsize{5.4}{6.5}\selectfont, text=black!70] at (9.28,\y) {M$_\i$};
  \foreach \j in {0,...,4}{
    \ifnum\j=\i\relax
      \draw[black!45, fill=white, line width=0.4pt] ({9.36+\j*0.20},{\y-0.09}) rectangle ++(0.16,0.18);
    \else
      \draw[clrCls!60!black, fill=clrCls!55, line width=0.35pt] ({9.36+\j*0.20},{\y-0.09}) rectangle ++(0.16,0.18);
    \fi
  }
  \draw[fine] (10.40,\y) -- (10.58,\y);
  \ifnum\keep=1\relax
    \node[chip, draw=clrEns!75!black, fill=clrEns!26, line width=0.8pt,
      minimum width=0.66cm] (s\i) at (10.92,\y) {$\fcv$};
    \draw[black!45, line width=0.5pt] (11.27,\y) -- (11.52,\y);
  \else
    \node[chip, draw=black!30, fill=black!5, text=black!40, minimum width=0.66cm]
      (s\i) at (10.92,\y) {$\fcv$};
  \fi
}
\draw[black!45, line width=0.5pt] (11.52,7.20) -- (11.52,4.92);
\draw[edge] (11.52,4.92) -- (10.27,4.92) -- (10.27,4.72);
\node[box, dashed, draw=clrEns!80!black, fill=clrEns!24, line width=0.8pt,
  minimum width=3.05cm, minimum height=0.52cm] at (10.27,4.44)
  {\textbf{branch} $=$ the three best $\fcv$};
\foreach \y/\clr/\ttl/\sub in {%
  7.62/clrCls/{Generalist (G)}/{trained on all subtasks},
  6.16/clrThird/{Specialist (S)}/{trained on one subtask},
  4.70/clrSFT/{G / S / MCS}/{submission-dependent}}{
  \node[rectangle, rounded corners=3pt, dashed, draw=\clr!70!black, fill=\clr!16,
    line width=0.8pt, minimum width=3.15cm, minimum height=1.08cm] at (14.02,\y) {};
  \node[font=\fontsize{5.8}{7}\selectfont\bfseries] at (14.02,{\y+0.30}) {\ttl};
  \node[ann, text=black!55] at (14.02,{\y+0.06}) {\sub};
  \foreach \dx in {-0.88, 0, 0.88}{
    \node[chip, draw=\clr!60!black, fill=white, minimum width=0.78cm,
      dash pattern=on 1.2pt off 1.2pt] at ({14.02+\dx},{\y-0.26}) {voter};
  }
}
\node[chip, draw=black!40, fill=black!4, minimum width=2.4cm] at (14.02,3.14)
  {$3$ branches $\times\,3$ folds $=$ \textbf{9} voters};
\draw[edge] (14.02,2.90) -- (14.02,2.34) -- (7.05,2.34) -- (7.05,2.10);
\node[rectangle, rounded corners=3pt, draw=clrEns!80!black, fill=clrEns!26,
  line width=0.9pt, minimum width=9.10cm, minimum height=0.74cm,
  font=\fontsize{6.2}{7.4}\selectfont] (vote) at (4.60,1.73)
  {$\widehat{Y}_t=P_t\!\left(\left\{\ell\in\mathcal{L}_t\mid
    \textstyle\sum_{j=1}^{9}\mathbf{1}[\ell\in v_j^{(t)}]\geq5\right\}\right),
    \quad t=1,2,3$};
\node[annl, text=black!62] at (0.05,2.26)
  {\textbf{majority vote}\quad five of nine};
\draw[edge] (9.15,1.73) -- (9.47,1.73);
\node[rectangle, rounded corners=3pt, draw=black!40, fill=white, line width=0.7pt,
  minimum width=3.48cm, minimum height=1.16cm] at (11.26,1.62) {};
\node[annl, text=black!62] at (9.62,2.02) {\textbf{taxonomy rules} $P_t$};
\node[annl, font=\fontsize{5.4}{6.5}\selectfont] at (9.62,1.76) {ST1 one class, tie $\rightarrow$ {\fontsize{4.8}{5.8}\selectfont\texttt{physical\_goods}}};
\node[annl, font=\fontsize{5.4}{6.5}\selectfont] at (9.62,1.50) {ST2 never empty, argmax as fallback};
\node[annl, font=\fontsize{5.4}{6.5}\selectfont] at (9.62,1.24) {ST3 exclusive flags $+$ disclosure XOR};
\draw[edge] (13.02,1.73) -- (13.32,1.73);
\node[rectangle, rounded corners=3pt, draw=black!45, fill=black!3, line width=0.7pt,
  minimum width=2.42cm, minimum height=1.16cm] at (14.58,1.62) {};
\node[annl, text=black!62] at (13.44,2.02) {\textbf{prediction}};
\node[annl, font=\fontsize{5.4}{6.5}\selectfont] at (13.44,1.76) {ST1 one commercial type};
\node[annl, font=\fontsize{5.4}{6.5}\selectfont] at (13.44,1.50) {ST2 product-category set};
\node[annl, font=\fontsize{5.4}{6.5}\selectfont] at (13.44,1.24) {ST3 compliance-flag set};
\node[rectangle, rounded corners=1.5pt, draw=black!55, dashed, line width=0.7pt,
  minimum width=0.42cm, minimum height=0.22cm] at (0.26,0.56) {};
\node[annl] at (0.52,0.56) {$=$ one branch};
\node[rectangle, rounded corners=1.5pt, draw=black!55,
  dash pattern=on 1.2pt off 1.2pt, line width=0.5pt,
  minimum width=0.42cm, minimum height=0.22cm] at (2.36,0.56) {};
\node[annl] at (2.62,0.56) {$=$ one trained model};
\draw[clrCls!60!black, fill=clrCls!55, line width=0.35pt] (5.10,0.48) rectangle ++(0.16,0.16);
\node[annl] at (5.36,0.56) {$=$ train fold};
\draw[black!45, fill=white, line width=0.4pt] (6.90,0.48) rectangle ++(0.16,0.16);
\node[annl] at (7.16,0.56) {$=$ held-out fold};
\fill[black!8] (8.95,0.48) rectangle ++(0.30,0.16);
\draw[black!32, line width=0.3pt] (8.95,0.48) -- (9.25,0.64);
\draw[black!32, line width=0.3pt] (8.95,0.64) -- (9.25,0.48);
\draw[black!40, line width=0.3pt] (8.95,0.48) rectangle ++(0.30,0.16);
\node[annl] at (9.35,0.56) {$=$ label mass an MCS never sees};
\end{tikzpicture}%
}
\caption{The system end to end, from the sponsored segment and its access level (a) through the voter pool (b) and cross-validation (c) to the deployed nine-voter ensemble (d).}
\label{fig:pipeline}
\end{figure*}
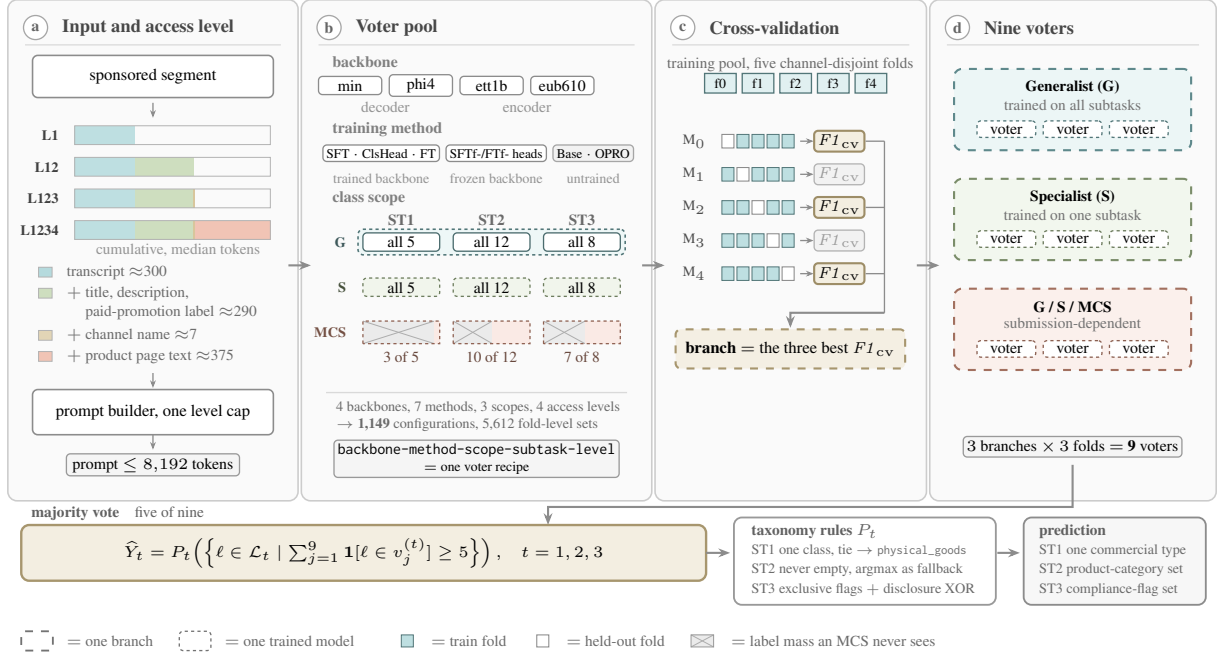

\section{The System End to End}
\label{app:pipeline}
Figure~\ref{fig:pipeline} draws the whole pipeline in one picture.
Panel~(a) is the input, a sponsored segment whose child-facing channel and commercial nature are given, with the four cumulative access levels and their median token counts.
One builder assembles the prompt of at most 8{,}192 tokens under a level cap (Appendix~\ref{app:prompts}).
Panel~(b) is the voter pool of \S\ref{sec:pool} and \S\ref{sec:scopes}, four backbones, seven training methods and three class scopes, and panel~(c) the cross-validation of \S\ref{sec:cv}.
Panel~(d) is the deployed ensemble, three branches of three fold-models, whose votes the rule of \S\ref{sec:voting} and the taxonomy rules of \S\ref{sec:rules} turn into one commercial type, a set of product categories and a set of compliance flags.
Panels~(b) and~(c) expand Figures~\ref{fig:scopes} and~\ref{fig:cv5}.

\section{Complete Access-Level Results}
\label{app:level-table}
\begin{figure*}[t]
\centering
\includegraphics[width=\textwidth]{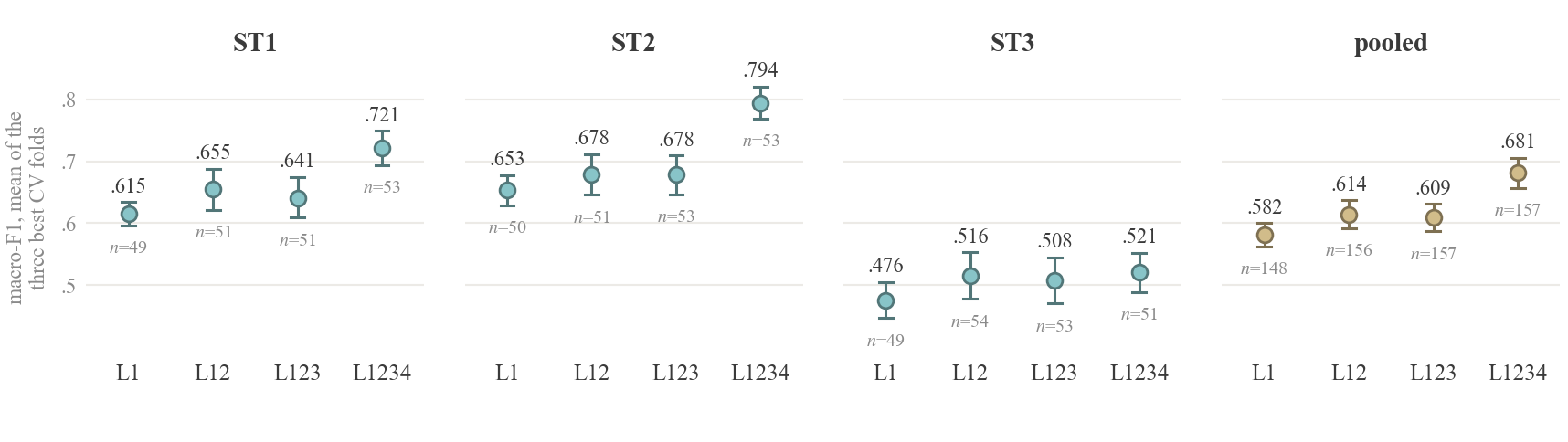}
\caption{Voter-level $\fcvtop$ by access level, per subtask and pooled. Whiskers are 95\% t-CIs and $n$ denotes observations.}
\label{fig:levels-app}
\end{figure*}

Figure~\ref{fig:levels-app} gives the task-specific values behind Table~\ref{tab:pooled-levels}.

\section{Trained and Untrained Methods}
\label{app:methods}
\begin{figure*}[t]
\centering
\includegraphics[width=\textwidth]{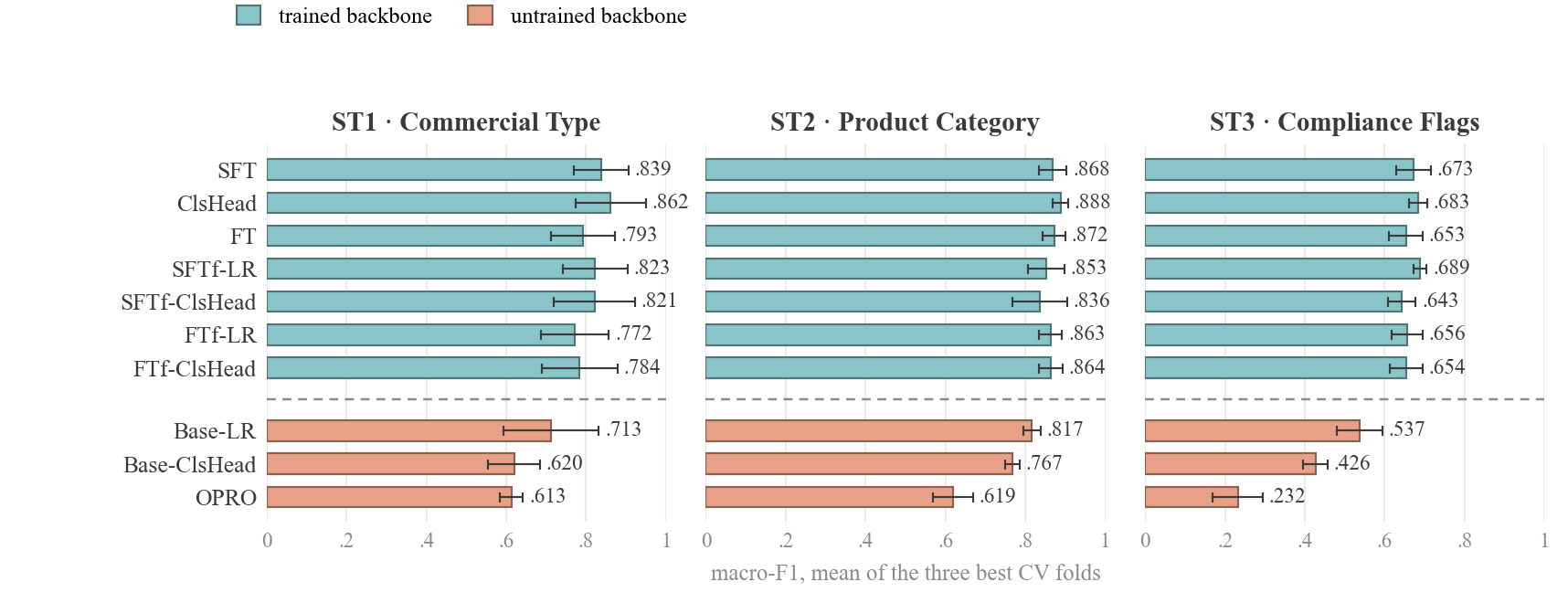}
\caption{Best $\fcvtop$ per training method and subtask at L1234 under train-pool CV5. Bars above the dashed line use a trained backbone, and whiskers span one standard deviation over the configuration's five fold scores.}
\label{fig:methods}
\end{figure*}

Figure~\ref{fig:methods} compares the best fully finished five-fold configuration of every training method at full access.
Every trained method beats every untrained one on every subtask, and OPRO's best ST3 across all levels is $0.373$ (L123).

\section{Prompts}
\label{app:prompts}
One builder assembles a system and a user message from a datapoint and a level cap.
Every voter reads their concatenation, encoders and frozen-backbone voters included, and training and inference call the same builder.

\begin{tcolorbox}[promptbox,
  title={\fontsize{8}{9.5}\selectfont\bfseries Voter prompt, assembled per datapoint and level cap},
]
\textbf{System:} You are a compliance analyst for EU consumer law (UCPD, AVMSD, DSA, CRD). \ldots{} Two facts are given and must not be re-assessed: the channel is child-facing, and the segment is commercial. \ldots{} You always answer with exactly one line in the format \texttt{labels: name, name, ...}\\[1.5pt]
\textbf{User:} Below is a sponsored segment from a YouTube video on a child-facing channel.\\[1.5pt]
{\ttfamily\raggedright
TRANSCRIPT: [transcript] \hfill (L1)\\
VIDEO TITLE: [title] \hfill (L2)\\
PLATFORM PAID-PROMOTION LABEL: yes|no|unknown \hfill (L2)\\
VIDEO DESCRIPTION: [description] \hfill (L2)\\
CHANNEL: [channel name] \hfill (L3)\\
PRODUCT PAGE ([page title]): [page text] \hfill (L4)\\[1.5pt]
\#\# Task: [subtask name]\\
\#\# Categories\par}
{\itshape [one definition per label, quoted from the official taxonomy]}\\[1.5pt]
Select one or more labels from: \texttt{[label inventory]}.
\end{tcolorbox}

The system message adds one decision rule per subtask, ST1 decides from what the buyer receives rather than from how the offer is marketed, ST2 notes that one offer often carries several categories, and ST3 names the instruments and the exclusive flags.
The category block quotes the official taxonomy verbatim, so the label definitions reach the model as text.
OPRO \citep{yang2023opro} instead searches its own system message per subtask, class scope and access level, scored on a class-balanced training sample and never on the development set or the scored fold, and a parser enforces the ST3 rules.

\section{Training and Thresholds}
\label{app:training}
\balance
All decoder voters train with 4-bit NF4 QLoRA on all linear projections for 10~epochs at effective batch size~8 and sequence length 8{,}192, and we deploy the checkpoint of the best held-out epoch.
SFT uses LoRA rank~32 ($\alpha{=}64$) at learning rate $10^{-4}$.
ClsHead uses rank~16 ($\alpha{=}32$) at $2{\times}10^{-5}$ with focal loss \citep{lin2017focal} and inverse-frequency class weights.
Encoders train all weights on the same schedule (10~epochs, learning rate $2{\times}10^{-5}$, effective batch~8) with the same heads and losses and mean pooling, and \texttt{ett1b}'s native 7{,}999-token window clips one product page at full access.
Frozen heads use a per-label L2 logistic regression with a per-fold strength sweep or a two-layer MLP (hidden${\rightarrow}$hidden$/2$, GELU).
These fits are seeded with 42 plus the fold index, while the QLoRA runs were not seed-pinned, and each ST1 fold-model uses argmax.
Multi-label thresholds are tuned per label on each fold-model's own held-out fold over the grid $\{1/40,\dots,39/40\}$, where two equal-scoring thresholds keep the higher one and a label without validation positives never fires.

\end{document}